\documentclass[sigconf,nonacm]{acmart}
\usepackage{tabularx}
\usepackage{gensymb}
\usepackage{tikz}
\usepackage{xcolor}
\usepackage{pgfplots}
\usepackage{siunitx}
\usepackage[inline]{enumitem}
\usepackage{cleveref}
\pgfplotsset{compat=1.18}

\usetikzlibrary{
    arrows,
    calc,
    positioning
}

\definecolor{kit-green}{RGB}{0, 150, 130}
\definecolor{kit-green100}{RGB}{0, 150, 130}
\definecolor{kit-green90}{rgb}{0.1, 0.6294, 0.5588}
\definecolor{kit-green80}{rgb}{0.2, 0.6706, 0.6078}
\definecolor{kit-green75}{rgb}{0.25, 0.6912, 0.6324}
\definecolor{kit-green70}{rgb}{0.3, 0.7118, 0.6569}
\definecolor{kit-green60}{rgb}{0.4, 0.7529, 0.7059}
\definecolor{kit-green50}{rgb}{0.5, 0.7941, 0.7549}
\definecolor{kit-green40}{rgb}{0.6, 0.8353, 0.8039}
\definecolor{kit-green30}{rgb}{0.7, 0.8765, 0.8529}
\definecolor{kit-green25}{rgb}{0.75, 0.8971, 0.8775}
\definecolor{kit-green20}{rgb}{0.8, 0.9176, 0.902}
\definecolor{kit-green15}{rgb}{0.85, 0.9382, 0.9265}
\definecolor{kit-green10}{rgb}{0.9, 0.9588, 0.951}
\definecolor{kit-green5}{rgb}{0.95, 0.9794, 0.9755}

\definecolor{kit-blue}{RGB}{70, 100, 170}
\definecolor{kit-blue100}{RGB}{70, 100, 170}
\definecolor{kit-blue90}{rgb}{0.3471, 0.4529, 0.7}
\definecolor{kit-blue80}{rgb}{0.4196, 0.5137, 0.7333}
\definecolor{kit-blue75}{rgb}{0.4559, 0.5441, 0.75}
\definecolor{kit-blue70}{rgb}{0.4922, 0.5745, 0.7667}
\definecolor{kit-blue60}{rgb}{0.5647, 0.6353, 0.8}
\definecolor{kit-blue50}{rgb}{0.6373, 0.6961, 0.8333}
\definecolor{kit-blue40}{rgb}{0.7098, 0.7569, 0.8667}
\definecolor{kit-blue30}{rgb}{0.7824, 0.8176, 0.9}
\definecolor{kit-blue25}{rgb}{0.8186, 0.848, 0.9167}
\definecolor{kit-blue20}{rgb}{0.8549, 0.8784, 0.9333}
\definecolor{kit-blue15}{rgb}{0.8912, 0.9088, 0.95}
\definecolor{kit-blue10}{rgb}{0.9275, 0.9392, 0.9667}
\definecolor{kit-blue5}{rgb}{0.9637, 0.9696, 0.9833}

\definecolor{kit-red}{RGB}{162, 34, 35}
\definecolor{kit-red100}{RGB}{162, 34, 35}
\definecolor{kit-red90}{rgb}{0.6718, 0.22, 0.2235}
\definecolor{kit-red80}{rgb}{0.7082, 0.3067, 0.3098}
\definecolor{kit-red75}{rgb}{0.7265, 0.35, 0.3529}
\definecolor{kit-red70}{rgb}{0.7447, 0.3933, 0.3961}
\definecolor{kit-red60}{rgb}{0.7812, 0.48, 0.4824}
\definecolor{kit-red50}{rgb}{0.8176, 0.5667, 0.5686}
\definecolor{kit-red40}{rgb}{0.8541, 0.6533, 0.6549}
\definecolor{kit-red30}{rgb}{0.8906, 0.74, 0.7412}
\definecolor{kit-red25}{rgb}{0.9088, 0.7833, 0.7843}
\definecolor{kit-red20}{rgb}{0.9271, 0.8267, 0.8275}
\definecolor{kit-red15}{rgb}{0.9453, 0.87, 0.8706}
\definecolor{kit-red10}{rgb}{0.9635, 0.9133, 0.9137}
\definecolor{kit-red5}{rgb}{0.9818, 0.9567, 0.9569}

\definecolor{kit-yellow}{RGB}{252, 229, 0}
\definecolor{kit-yellow100}{RGB}{252, 229, 0}
\definecolor{kit-yellow90}{rgb}{0.9894, 0.9082, 0.1}
\definecolor{kit-yellow80}{rgb}{0.9906, 0.9184, 0.2}
\definecolor{kit-yellow75}{rgb}{0.9912, 0.9235, 0.25}
\definecolor{kit-yellow70}{rgb}{0.9918, 0.9286, 0.3}
\definecolor{kit-yellow60}{rgb}{0.9929, 0.9388, 0.4}
\definecolor{kit-yellow50}{rgb}{0.9941, 0.949, 0.5}
\definecolor{kit-yellow40}{rgb}{0.9953, 0.9592, 0.6}
\definecolor{kit-yellow30}{rgb}{0.9965, 0.9694, 0.7}
\definecolor{kit-yellow25}{rgb}{0.9971, 0.9745, 0.75}
\definecolor{kit-yellow20}{rgb}{0.9976, 0.9796, 0.8}
\definecolor{kit-yellow15}{rgb}{0.9982, 0.9847, 0.85}
\definecolor{kit-yellow10}{rgb}{0.9988, 0.9898, 0.9}
\definecolor{kit-yellow5}{rgb}{0.9994, 0.9949, 0.95}

\definecolor{kit-orange}{RGB}{223, 155, 27}
\definecolor{kit-orange100}{RGB}{223, 155, 27}
\definecolor{kit-orange90}{rgb}{0.8871, 0.6471, 0.1953}
\definecolor{kit-orange80}{rgb}{0.8996, 0.6863, 0.2847}
\definecolor{kit-orange75}{rgb}{0.9059, 0.7059, 0.3294}
\definecolor{kit-orange70}{rgb}{0.9122, 0.7255, 0.3741}
\definecolor{kit-orange60}{rgb}{0.9247, 0.7647, 0.4635}
\definecolor{kit-orange50}{rgb}{0.9373, 0.8039, 0.5529}
\definecolor{kit-orange40}{rgb}{0.9498, 0.8431, 0.6424}
\definecolor{kit-orange30}{rgb}{0.9624, 0.8824, 0.7318}
\definecolor{kit-orange25}{rgb}{0.9686, 0.902, 0.7765}
\definecolor{kit-orange20}{rgb}{0.9749, 0.9216, 0.8212}
\definecolor{kit-orange15}{rgb}{0.9812, 0.9412, 0.8659}
\definecolor{kit-orange10}{rgb}{0.9875, 0.9608, 0.9106}
\definecolor{kit-orange5}{rgb}{0.9937, 0.9804, 0.9553}

\definecolor{kit-lightgreen}{RGB}{140, 182, 60}
\definecolor{kit-lightgreen100}{RGB}{140, 182, 60}
\definecolor{kit-lightgreen90}{rgb}{0.5941, 0.7424, 0.3118}
\definecolor{kit-lightgreen80}{rgb}{0.6392, 0.771, 0.3882}
\definecolor{kit-lightgreen75}{rgb}{0.6618, 0.7853, 0.4265}
\definecolor{kit-lightgreen70}{rgb}{0.6843, 0.7996, 0.4647}
\definecolor{kit-lightgreen60}{rgb}{0.7294, 0.8282, 0.5412}
\definecolor{kit-lightgreen50}{rgb}{0.7745, 0.8569, 0.6176}
\definecolor{kit-lightgreen40}{rgb}{0.8196, 0.8855, 0.6941}
\definecolor{kit-lightgreen30}{rgb}{0.8647, 0.9141, 0.7706}
\definecolor{kit-lightgreen25}{rgb}{0.8873, 0.9284, 0.8088}
\definecolor{kit-lightgreen20}{rgb}{0.9098, 0.9427, 0.8471}
\definecolor{kit-lightgreen15}{rgb}{0.9324, 0.9571, 0.8853}
\definecolor{kit-lightgreen10}{rgb}{0.9549, 0.9714, 0.9235}
\definecolor{kit-lightgreen5}{rgb}{0.9775, 0.9857, 0.9618}

\definecolor{kit-purple}{RGB}{163, 16, 124}
\definecolor{kit-purple100}{RGB}{163, 16, 124}
\definecolor{kit-purple90}{rgb}{0.6753, 0.1565, 0.5376}
\definecolor{kit-purple80}{rgb}{0.7114, 0.2502, 0.589}
\definecolor{kit-purple75}{rgb}{0.7294, 0.2971, 0.6147}
\definecolor{kit-purple70}{rgb}{0.7475, 0.3439, 0.6404}
\definecolor{kit-purple60}{rgb}{0.7835, 0.4376, 0.6918}
\definecolor{kit-purple50}{rgb}{0.8196, 0.5314, 0.7431}
\definecolor{kit-purple40}{rgb}{0.8557, 0.6251, 0.7945}
\definecolor{kit-purple30}{rgb}{0.8918, 0.7188, 0.8459}
\definecolor{kit-purple25}{rgb}{0.9098, 0.7657, 0.8716}
\definecolor{kit-purple20}{rgb}{0.9278, 0.8125, 0.8973}
\definecolor{kit-purple15}{rgb}{0.9459, 0.8594, 0.9229}
\definecolor{kit-purple10}{rgb}{0.9639, 0.9063, 0.9486}
\definecolor{kit-purple5}{rgb}{0.982, 0.9531, 0.9743}

\definecolor{kit-brown}{RGB}{167, 130, 46}
\definecolor{kit-brown100}{RGB}{167, 130, 46}
\definecolor{kit-brown90}{rgb}{0.6894, 0.5588, 0.2624}
\definecolor{kit-brown80}{rgb}{0.7239, 0.6078, 0.3443}
\definecolor{kit-brown75}{rgb}{0.7412, 0.6324, 0.3853}
\definecolor{kit-brown70}{rgb}{0.7584, 0.6569, 0.4263}
\definecolor{kit-brown60}{rgb}{0.7929, 0.7059, 0.5082}
\definecolor{kit-brown50}{rgb}{0.8275, 0.7549, 0.5902}
\definecolor{kit-brown40}{rgb}{0.862, 0.8039, 0.6722}
\definecolor{kit-brown30}{rgb}{0.8965, 0.8529, 0.7541}
\definecolor{kit-brown25}{rgb}{0.9137, 0.8775, 0.7951}
\definecolor{kit-brown20}{rgb}{0.931, 0.902, 0.8361}
\definecolor{kit-brown15}{rgb}{0.9482, 0.9265, 0.8771}
\definecolor{kit-brown10}{rgb}{0.9655, 0.951, 0.918}
\definecolor{kit-brown5}{rgb}{0.9827, 0.9755, 0.959}

\definecolor{kit-cyan}{RGB}{35, 161, 224}
\definecolor{kit-cyan100}{RGB}{35, 161, 224}
\definecolor{kit-cyan90}{rgb}{0.2235, 0.6682, 0.8906}
\definecolor{kit-cyan80}{rgb}{0.3098, 0.7051, 0.9027}
\definecolor{kit-cyan75}{rgb}{0.3529, 0.7235, 0.9088}
\definecolor{kit-cyan70}{rgb}{0.3961, 0.742, 0.9149}
\definecolor{kit-cyan60}{rgb}{0.4824, 0.7788, 0.9271}
\definecolor{kit-cyan50}{rgb}{0.5686, 0.8157, 0.9392}
\definecolor{kit-cyan40}{rgb}{0.6549, 0.8525, 0.9514}
\definecolor{kit-cyan30}{rgb}{0.7412, 0.8894, 0.9635}
\definecolor{kit-cyan25}{rgb}{0.7843, 0.9078, 0.9696}
\definecolor{kit-cyan20}{rgb}{0.8275, 0.9263, 0.9757}
\definecolor{kit-cyan15}{rgb}{0.8706, 0.9447, 0.9818}
\definecolor{kit-cyan10}{rgb}{0.9137, 0.9631, 0.9878}
\definecolor{kit-cyan5}{rgb}{0.9569, 0.9816, 0.9939}

\definecolor{kit-gray}{RGB}{0, 0, 0}
\definecolor{kit-gray100}{RGB}{0, 0, 0}
\definecolor{kit-gray90}{rgb}{0.1, 0.1, 0.1}
\definecolor{kit-gray80}{rgb}{0.2, 0.2, 0.2}
\definecolor{kit-gray75}{rgb}{0.25, 0.25, 0.25}
\definecolor{kit-gray70}{rgb}{0.3, 0.3, 0.3}
\definecolor{kit-gray60}{rgb}{0.4, 0.4, 0.4}
\definecolor{kit-gray50}{rgb}{0.5, 0.5, 0.5}
\definecolor{kit-gray40}{rgb}{0.6, 0.6, 0.6}
\definecolor{kit-gray30}{rgb}{0.7, 0.7, 0.7}
\definecolor{kit-gray25}{rgb}{0.75, 0.75, 0.75}
\definecolor{kit-gray20}{rgb}{0.8, 0.8, 0.8}
\definecolor{kit-gray15}{rgb}{0.85, 0.85, 0.85}
\definecolor{kit-gray10}{rgb}{0.9, 0.9, 0.9}
\definecolor{kit-gray5}{rgb}{0.95, 0.95, 0.95}

\DeclareSIUnit\coe{CO_2e}
\DeclareSIUnit\gcoe{\gram\coe}
\DeclareSIUnit\gcoee{\gcoe\per\kilo\watt\per\hour}

\AtBeginDocument{%
  }

\acmConference[]{Technical Report}{June 25}{2025}

\begin{document}

\title[Jigsaw]{Jigsaw: Training Multi-Billion-Parameter AI Weather Models With Optimized Model Parallelism}

\author{Deifilia Kieckhefen}

\orcid{0009-0000-1197-5493} 
\email{deifilia.kieckhefen@kit.edu}
\affiliation{%
  \institution{Karlsruhe Institute of Technology}
  \city{Eggenstein-Leopoldshafen}
  \state{Baden-Württemberg}
  \country{Germany}
}
\author{Markus Götz}
\orcid{0000-0002-2233-1041} 
\email{markus.goetz@kit.edu}
\affiliation{%
  \institution{Karlsruhe Institute of Technology, Helmholtz AI}
  \city{Eggenstein-Leopoldshafen}
  \postcode{76344}
  \state{Baden-Württemberg}
  \country{Germany}
}
\author{Lars H. Heyen}
\orcid{0000-0001-7949-1858} 
\email{lars.heyen@kit.edu} 
\affiliation{%
  \institution{Karlsruhe Institute of Technology}
  \city{Eggenstein-Leopoldshafen}
  \postcode{76344}
  \state{Baden-Württemberg}
  \country{Germany}
}
\author{Achim Streit}
\orcid{0000-0002-5065-469X} 
\email{achim.streit@kit.edu}
\affiliation{%
  \institution{Karlsruhe Institute of Technology}
  \city{Eggenstein-Leopoldshafen}
  \postcode{76344}
  \state{Baden-Württemberg}
  \country{Germany}
}
\author{Charlotte Debus}
\orcid{0000-0002-7156-2022} 
\email{charlotte.debus@kit.edu}
\affiliation{%
  \institution{Karlsruhe Institute of Technology}
  \city{Eggenstein-Leopoldshafen}
  \postcode{76344}
  \state{Baden-Württemberg}
  \country{Germany}
}

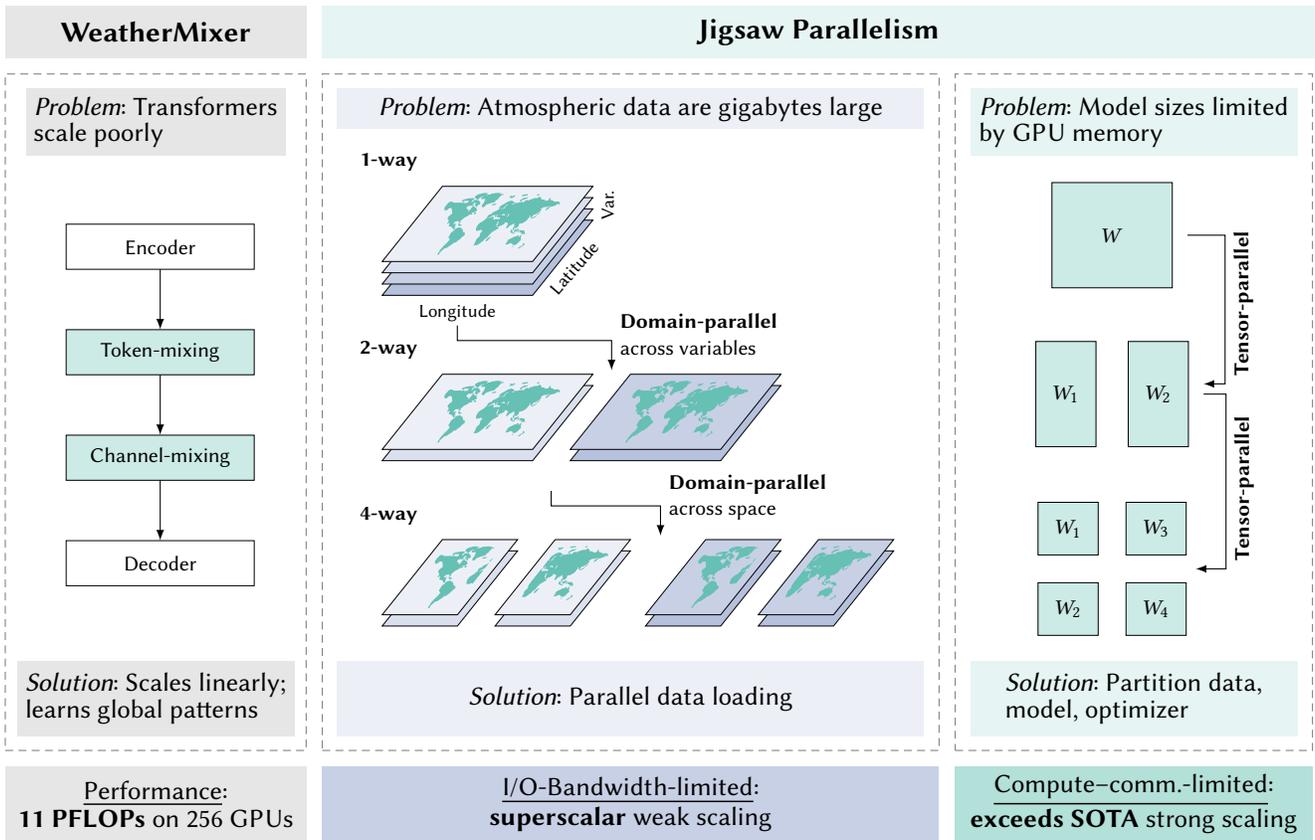
\renewcommand{\shortauthors}{Kieckhefen et al.}

\begin{abstract} 
AI-based methods have revolutionized atmospheric forecasting, with recent successes in medium-range forecasting spurring the development of climate foundation models. Accurate modeling of complex atmospheric dynamics at high spatial resolutions and longer lead times requires large neural networks and gigabyte-sized data samples, making accelerator memory and I/O-bandwidth the bottlenecks for model training.
We introduce WeatherMixer, a multi-layer-perceptron-based architecture whose workload scales linearly with input size, allowing the model to learn global weather phenomena at accuracies similar to numerical weather prediction. To cope with the computational demand, we propose Jigsaw, a novel model parallelization scheme that employs both domain and tensor parallelism, eliminating memory redundancy. Jigsaw exceeds state-of-the-art performance in strong scaling in compute--communication-limited systems and achieves superscalar weak scaling in I/O-bandwidth-limited systems. We scale training to 256 GPUs, reaching peak performances of 9 and 11 PFLOPs---23\% and 28\% of theoretical peaks---achieving 68\% and 72\% scaling efficiency versus 51\% without model parallelism.

\end{abstract}



\begin{teaserfigure}
  \input{figures/graphical_abstract.tikz}
  \Description{Visual depiction of Jigsaw parallelism scheme and WeatherMixer.}
  \label{fig:teaser}
\end{teaserfigure}


\maketitle
\setlength{\belowcaptionskip}{-10pt}
\section{Introduction}
The recent emergence of accurate, artificial intelligence (AI)-based atmospheric forecast models is a revolution in meteorology. 
Historically, this discipline has been dominated by process-based numerical modeling. Practically, this entails solving partial differential equations that describe atmospheric dynamics with numerical methods. 
Data-driven models instead learn purely from historic data without requiring explicit formulations of the underlying physical equations.
The predictive skill of data-driven models now exceeds that of classical numerical approaches in many of the standard forecast scores \cite{keisler2022forecastingglobalweathergraph,Lam2023,Bi_2023,nguyen2024scalingtransformerneuralnetworks}.
Almost all of the recent breakthroughs in data-driven weather forecasting and initial attempts towards atmospheric foundation models are based on the Transformer neural architecture~\cite{vaswani2017attention}. The predictive capabilities of the attention mechanism have pushed data-driven models from an interesting research question towards the possibility of operational deployment.
Current attempts at further improving forecasting skill ~\cite{wang2024,bodnar2024aurora} generally try to exploit neural network scaling laws~\cite{kaplan2020scaling}, i.e., an increase in the number of free trainable parameters in the model and a larger training dataset size should result in better generalizability and predictive performance. However, the resulting models have immense sizes. In combination with the large size of the input data, the memory capacity of individual accelerators (GPUs) is rapidly exceeded. To overcome this, different parallelization and distributed-memory computation approaches have been pursued. Next to the widely employed data parallelism, i.e., distributing the dataset into disjoint batches across computation processes and training local model copies on these data chunks, more recent works have turned to model parallelism (MP), i.e., distributing the weights of the models themselves. 
Approaches like Megatron-LM~\cite{shoeybi2020}, ZeRO~\cite{rajbhandari2020}, and FSDP~\cite{zhao2023}, originally developed for large language models (LLMs), do not tackle scalability limitations of Transformers with respect to global atmospheric data well.
For one, the individual input samples in atmospheric sciences are orders of magnitude larger in size than that of LLMs, due to the highly-resolved image-like data with tens or even hundreds of individual state variables (channels). Transformer-based atmospheric models commonly employ tokenization in the spatial domain, thus suffering from a quadratic computational complexity of the self-attention mechanism. While current models are still able to handle the global 0.25\degree{} resolution of the commonly used ECMWF Reanalysis v5  dataset (ERA5)~\cite{Hersbach2020} by training with only one or two data items per accelerator at a time, higher spatial resolutions, as to be expected for ERA6 or CMIP6, will inevitably exceed the hardware capabilities.

Second, for true operational deployment, high predictive performance for longer lead times in the three- to seven-day medium-range forecast time is required. In this temporal range, atmospheric phenomena at global spatial scales become important. Thus, ``global vision'' of the model is necessary---this is currently infeasible for current modeling approaches and operational compute budgets. Current Transformer-based models are limited to computing highly localized spatial extents, e.g., via shifted window attention~\cite{liu2021}. 
Similarly, to achieve a sufficient spatial interaction range in graph-based neural models, the network's depth needs to be increased linearly. This inherently implies increasing model size and, correspondingly, computational load. 

In response to these challenges for large data-driven meteorological models---higher spatial resolutions and longer rollout times---we propose WeatherMixer, an atmospheric model based on the MLP-Mixer architecture~\cite{tolstikhin2021}.
Multi-layer perceptions (MLPs) have experienced a revival, as a number of studies demonstrated their competitive predictive skills for time-series forecasting~\cite{Ekambaram_2023,chen2023tsmixerallmlparchitecturetime}. From a computational perspective, this has the major advantage that such feed-forward models reduce to a sequence of matrix--matrix multiplications, as opposed to the complex computational pathway of the self-attention mechanism. As such, WeatherMixer provides a framework for model parallelism across all network layers via distributed dense matrix--matrix multiplications. In this work, we extend upon this principle with Jigsaw parallelism, which provides efficient and scalable intra-node distribution of the WeatherMixer towards multi-billion-parameter model capacities.

Our main contributions are WeatherMixer, a new MLP-based architecture for atmospheric dynamics and climate forecasting; and Jigsaw, a novel parallelism approach that fully shards data, model, and optimizer states. 
Jigsaw partitions model weights and training data across one dimension when distributed over two compute devices and two dimensions when distributed over four compute devices within a node.
For the first time in a data-driven atmospheric model, we present roofline plots that show the I/O- and computation--communication-bounds when working with different precisions. We present strong scaling results that rival state-of-the-art LLM-parallelism approaches and show that Jigsaw achieves superscalar weak scaling in I/O-bandwidth-limited systems. To quantify the resources consumed for training, we also report the measured energy consumption and estimate carbon-emissions footprint for our primary training experiments.

\section{Related Work}

\subsection{Data-driven atmospheric models}

Although purely data-driven atmospheric models have been explored for some time~\cite{kurth2018exascale, dueben2018challenges, rasp2021data}, only recent developments, occurring over the past three years, have lead to fundamental breakthroughs. Among the first models to be published was
FourCastNet~\cite{pathak2022}. Utilizing adaptive Fourier neural operators, it is able to predict twenty atmospheric state variables at 0.25\degree{} global resolution with high accuracy. Global weather patterns are modeled via a Fourier-transform-based token mixing scheme on top of a vision Transformer~\cite{dosovitskiy2020image} backbone. 
Pangu-Weather, operating at the same resolution, was published shortly after~\cite{Bi_2023} and is considered to be the first model to exceed the performance of operational numerical weather predictions (the Integrated Forecasting System (IFS)). Pangu-Weather is based on a three-dimensional, shifted window Transformer~\cite{liu2021} with an Earth-specific bias term. The model's spatial interaction range is highly restricted by the memory requirements of self-attention performed over three-dimensional volumes.
More recently, GraphCast~\cite{Lam2023}, a graph-based model, uses a hierarchical multi-mesh to capture global atmospheric phenomena. Fine-tuned to long rollout times of 10 days, it surpasses the Transformer-based models in forecast skill on long lead times. However, higher spatial resolutions
can only be incorporated through more finely resolved multi-meshes, ultimately limiting  scalability due to available memory.

Since the introduction of these three seminal models, the computational requirements of training such atmospheric models have only increased, as higher-resolution regional models~\cite{nipen2024} and longer rollouts~\cite{boris2023} for medium-range or subseasonal forecasting~\cite{Lam2023,chen2023fuxicascademachinelearning} have become the next goals. 
In parallel, data-driven atmospheric foundation models have spurred significant research interest. These large, general-purpose models are pre-trained in a self-supervised manner on huge, heterogeneous datasets and can subsequently be fine-tuned to specific downstream tasks on smaller, specialized datasets.
Since the introduction of ClimaX~\cite{nguyen2023climax}, the first atmospheric foundation model, several successful approaches have been presented~\cite{bodnar2024aurora,lessig2023,wang2024}. Aurora~\cite{bodnar2024aurora} is a foundation model trained on a diverse range of datasets, including ERA5~\cite{Hersbach2020}, HRES forecasts and analyses~\cite{ifs78758}, IFS ENS~\cite{ifs78758}, CAMS~\cite{acp-19-3515-2019}, and CMIP6~\cite{CMPI5}. To scale the model within hardware constraints, it employs BFloat16 mixed precision, gradient checkpointing, and model parallelism via sharding of weights and gradients across multiple GPUs. For fine-tuning on longer rollouts at 0.1\degree{} global resolution, Low-Rank Adaptation (LoRA) layers~\cite{hu2022lora} are incorporated.

\begin{figure*}[tb]
    \centering
    \input{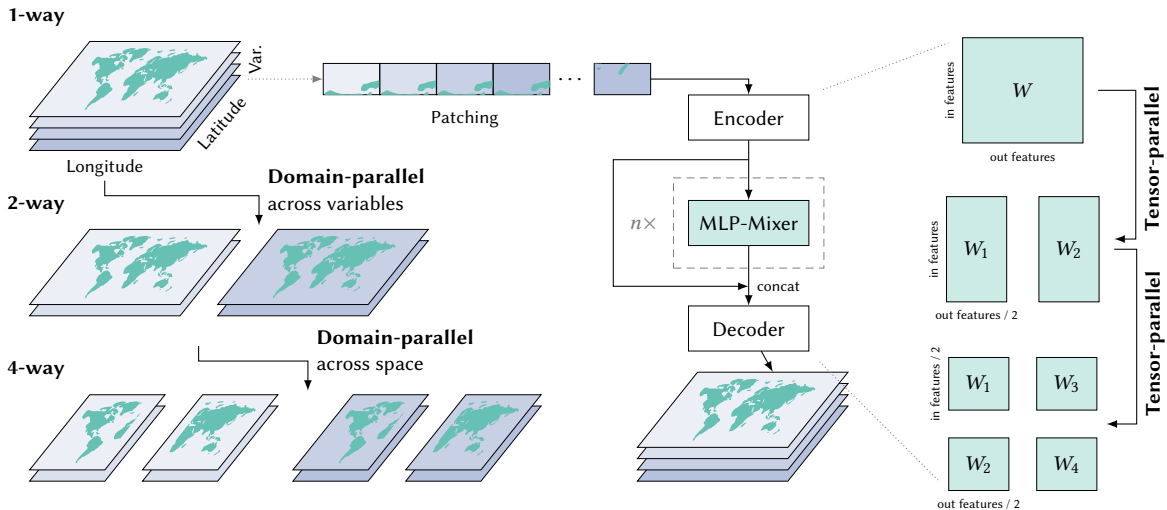}
    \caption{Jigsaw parallelism applied to WeatherMixer model. Jigsaw applies domain parallelism over individual data samples, allowing each model-parallel process to load non-overlapping segments of the data. Tensor parallelism is realized by distributing the model weights over the different processes.}
    \label{fig:jigsaw-parallelism}
    \Description{This needs a description}
\end{figure*}

\subsection{Parallelization schemes}

\subsubsection{Data parallelism}

The predictive capabilities of data-driven models are directly coupled to the amount of training data~\cite{kaplan2020scaling}. Increasing dataset sizes also require more computations, and training on a single computational device quickly becomes infeasible. Data-parallel (DP) training is a well-established technique to decrease the training time inversely proportional to the number of processors~\cite{ben2019demystifying}. In DP training, each process holds an identical copy of the model weights, but a disjoint subset of the entire data. Each process independently performs a forward--backward pass, computing local gradients. These are then averaged and updated across all processes in an allreduce operation. This synchronization point is the algorithm's main communication bottleneck. To improve scalability, overlapping communication with the backward pass~\cite{yamazaki2019yet} as well as asynchronous and hierarchical updates utilizing stale gradients have been explored~\cite{DeSa2017AsyncSGD,coquelin2022accelerating}. Moreover, data-parallel training is limited by a phenomenon called large batch effects~\cite{keskar2016large}. Higher degrees of parallelization imply larger effective global batch sizes, which can lead to poor model convergence and generalization performance.

\subsubsection{Model parallelism}
Next to the amount of training data, model size is the other key factor influencing predictive performance~\cite{kaplan2020scaling}. With growing model and data size, training becomes bound by the memory capacity, rather than the computational capabilities, of existing accelerator hardware. Since data samples are typically fixed in size, the number of free model parameters is the main factor of the memory footprint. Model parallelism has emerged as a tool for training larger models exceeding the memory capacity of a single accelerator by distributing parameters and/or optimizer states across multiple devices.
Megatron-LM, for example, trains multi-billion-parameter LLMs~\cite{shoeybi2020} by parallelizing across self-attention heads and feed-forward weight matrices in Transformer architectures. Feed-forward layers are parallelized in pairs: the first linear layer's weights are distributed along the output dimension, and the second linear layer's weights are distributed across the input dimension. This way, Megatron-LM reduces the number of synchronization points to a single allreduce operation in the forward pass and another in the backward pass.
The authors reported extensive scaling experiments, which serve as the reference baseline for our work. 
ZeRO~\cite{rajbhandari2020} improves upon Megatron-LM and builds 100-billion-parameter LLMs, performing extensive analysis on communication volume and memory requirements. They partition the parameters, but also gradient and optimizer states, achieving superlinear scaling due to reduced memory footprints. However, their work focuses on the analysis of memory consumption requirements, rather than scaling performance.
Fully Sharded Data Parallelism (FSDP) shards data batches and model parameters~\cite{zhao2023}. In a given layer in the forward pass, the weights are gathered to compute the full activations. Afterwards, the unnecessary activations are discarded to reduce the memory footprint on individual GPUs.

Data-driven atmospheric forecasting has yet to catch up with LLMs with respect to model parallelism: the first works are starting to explore corresponding algorithmic strategies \cite{Kurth2023,bodnar2024aurora,lessig2023}. Many models seek to increase the size of the model by employing techniques that will trade off increasing computational time for decreased memory consumption. The simplest method is to decrease the local batch size to one sample per GPU. Models also use mixed precision and autocasting so that the gradients and optimizer states require less memory~\cite{bodnar2024aurora}. Gradient checkpointing, in which gradients are recomputed when necessary in the backward pass instead of being stored in memory, can substantially reduce required memory at the cost of increased computational time. In addition, a few works have employed MP to train models that are larger than can fit on one GPU.
FourCastNet~\cite{Kurth2023} employs model parallelism by distributing embarrassingly parallel MLPs that learn on different channels. They also parallelize the matrix--matrix multiplication in the MLP by exploiting the fixed ratio of hidden/embedding dimension and implementing the matrix multiplications in the MLP point-wise convolutional operations. 
AIFS implements tensor parallelism, sharding processor and attention heads across multiple GPUs~\cite{lang2024}.  
Both AIFS and FourCastNet train on global 0.25\degree{} resolution data.
AtmoRep has a Multiformer architecture and trains a model parallelized across four GPUs within a node by distributing state variables to different compute devices.
ORBIT~\cite{wang2024} is the first work in this field that details their parallelization mechanism and scales beyond the 1-billion-parameter model size by combining various parallelization techniques. The architecture of ORBIT is a vision Transformer, and it is trained on global 1.4\degree{} resolution with up to 91 channels, i.e., 10 MB of data per sample in single precision. ORBIT introduces Hybrid-STOP, which fully shards the model's weights as well as data batches, thus combining tensor parallelism and FSDP. Additionally, communication is optimized by using layer wrapping, i.e., overlapping computation and prefetching of subsequent model layers. Through these means, they train a model with up to 113 billion parameters across 49,152 GPUs.


Common among all of the listed parallelization techniques is domain parallelism is not employed. Instead, entire data samples are loaded onto each model-parallel process, and only the model weights and optimizer states are sharded. Many parallelization schemes also require allgather operations to collect and further redistribute the weight tensors on individual GPUs~\cite{shoeybi2020,wang2024,zhao2023}.

\section{WeatherMixer Architecture}

\begin{figure*}
    \centering
    \input{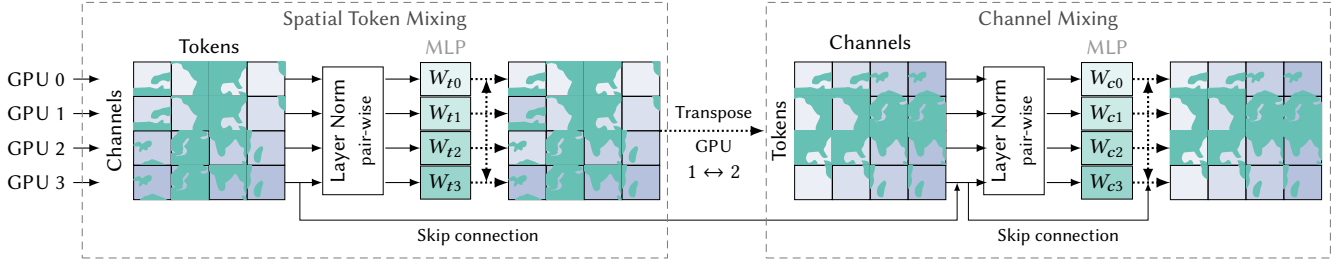}
    \caption{4-way MLP-Mixer schematic. The MLP-Mixer consists of two MLPs: First, an MLP is  applied across the spatial tokens, then a second MLP is applied across channels.}
    \label{fig:mlpmixer}
    \Description{This needs a description}
\end{figure*}

We introduce WeatherMixer (WM), inspired by the MLP-Mixer developed for image classification~\cite{tolstikhin2021}. A visual schematic of the architecture is depicted in Figure~\ref{fig:jigsaw-parallelism}, with the mixing blocks detailed in Figure~\ref{fig:mlpmixer}. 
The WM architecture features convolutional layers and MLP layers in an encoder--processor--decoder structure.

We consider an individual data sample to be a three-dimensional tensor, with two spatial dimensions and the atmospheric state variables in the third dimension. The encoder learns an embedding from the physical variables to a latent space of dimension $d_\mathrm{emb}$ and is implemented with a convolutional layer over non-overlapping spatial shifted windows. 
This compresses the input data into a sequence of spatial patches (tokens) in the latent space (channels). 

The processor consists of a series of mixing blocks. 
Each mixing block comprises two MLPs: one for token mixing and one for channel mixing.
For token mixing, the input tensor of dimensions [tokens, channels] is first transposed, and an MLP is independently applied across all of the tokens for each channel to learn the spatial relationships of individual atmospheric state variables. We emphasize that this construction allows for global weather patterns to be learned. 
Then, the data tensor is transposed again, and another MLP is applied across all of the channels for each spatial token. This channel mixing allows the model to learn the relationship between different atmospheric variables. 
Each token mixing and channel mixing MLP consists of two linear layers and a non-linear GELU activation function with optional dropout. 
Layer norms and residual skip connections are applied before and around each MLP.
To recover the physical atmospheric variables from the tokens and channels, a convolutional layer---the decoder---is computed. 
A forecast is obtained by calculating the weighted fraction between the input data and the model output, by means of a final linear layer.

\begin{figure}
    \centering
    \begin{tikzpicture}
    \sffamily
    \begin{axis}[
        axis on top,
        width=0.7\linewidth,
        height=0.6\linewidth,
        xlabel=Epoch,
        ylabel=Validation loss,
        label style={font=\footnotesize\bfseries},
        tick label style={
            font=\footnotesize\bfseries\sffamily,
            /pgf/number format/1000 sep=
        },
        axis x line=bottom,
        axis y line=left,
        xmode=linear,
        ymode=linear,
        xmin=0,
        xmax=100,
        ymin=0.01,
        ymax=0.1,
        grid=major,
        grid style={ 
            draw=kit-gray30,
            densely dotted
        },
        legend style={
            draw=none,
            fill=none,
            font=\footnotesize\sffamily,
            at={(1,1)},
            anchor=north east,
            nodes={scale=0.8, transform shape}
        },
        legend cell align={left}
    ]
        \addplot[
            color=kit-green, 
            point meta=explicit symbolic, 
            nodes near coords,
            nodes near coords align={
                yshift=6.3
            },
            every node near coord/.style={
                font=\tiny,
                color=black
            },
        ] table [x=epoch, y=model_0, col sep=comma] {figures/data/val_losses.csv};
        \addlegendentry{250 million parameters}
        \addplot[
            color=kit-blue, 
            point meta=explicit symbolic, 
            nodes near coords,
            nodes near coords align={
                yshift=6.3
            },
            every node near coord/.style={
                font=\tiny,
                color=black
            },
        ] table [x=epoch, y=model_1, col sep=comma] {figures/data/val_losses.csv};
        \addlegendentry{500 million parameters}
        \addplot[
            color=kit-orange, 
            point meta=explicit symbolic, 
            nodes near coords,
            nodes near coords align={
                yshift=6.3
            },
            every node near coord/.style={
                font=\tiny,
                color=black
            },
        ] table [x=epoch, y=model_2, col sep=comma] {figures/data/val_losses.csv};
        \addlegendentry{1 billion parameters}
    \end{axis}
\end{tikzpicture}
    \caption{Validation loss of three model sizes, trained on subsampled ERA5 6h-data from 1979--2017 and validated on 2018 data.}
    \label{fig:bigger_is_better}
    \Description{Placeholder description}
\end{figure}
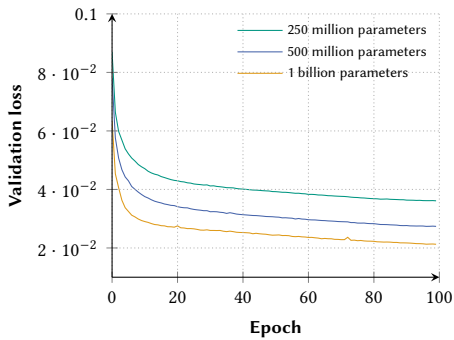

In line with neural scaling laws, we find that larger models, i.e., more free parameters, yield better predictive performance. Figure~\ref{fig:bigger_is_better} shows the validation loss of three increasingly large WeatherMixer models, with 250 million, 500 million, and 1 billion parameters, respectively. Details on the training setup and experimental evaluation can be found in \Cref{subec:model-performance}.
Typically, model size is limited by the available hardware memory, in particular for the large data samples in atmospheric science.

\section{Jigsaw parallelism}

Jigsaw parallelism is a distributed matrix--matrix multiplication method optimized for model parallel training of models with large input data. The model inputs are a four-dimensional tensor with dimensions [batch size, latitude, longitude, channels]. Given data $X$ and weights $W$, we follow the common convention of representing a linear layer via the matrix--matrix multiplication $XW^T$.

Jigsaw distributes not only the model parameter and optimizer states, i.e., tensor parallelism, but also employs domain parallelism by sharding the input sample. We define an $n$-way parallel model as one in which each GPU holds $1/n$ of the total parameters, optimizer states, and input sample at a given time. No allgather or broadcast communication is required during a forward--backward pass to collect all parameters on a single process. Thus, this parallelization scheme has \emph{zero memory redundancy} (aside from necessary buffers for communication). Jigsaw also lends itself well to being combined with other classical forms of parallelism, such as data and pipeline.

\subsection{Two-way parallelism}
In 2-way parallelism, we partition the model across two GPUs by splitting the data and parameter tensors across the final dimension---in this case, the channel dimension.
Given two ranks, 0 and 1, the global data $X$ and weights $W$ are thus distributed as $X = [X_0, X_1]$ and $W = [W_0, W_1]$. To maintain this partitioning scheme after the matrix--matrix multiplication $XW^T$, each process internally splits its data and weights along the second-to-last dimension, i.e., $X_0 = \begin{bmatrix}
    X_{0,0} &
    X_{0,1}
\end{bmatrix}^T$ and $ X_1 = \begin{bmatrix}
    X_{1,0} &
    X_{1,1}
\end{bmatrix}^T$.
The forward pass of a linear layer can be represented as:
\begin{align}
    XW^T &= \begin{bmatrix}
        X_{0,0} & X_{1,0} \\
        X_{0,1} & X_{1,1}
    \end{bmatrix}
    \begin{bmatrix}
        W_{0,0}^T & W_{0,1}^T \\
        W_{1,0}^T & W_{1,1}^T \\
    \end{bmatrix} \\
    &=
    \begin{bmatrix}
        X_{0,0}W_{0,0}^T + \mathbf{X_{1,0}W_{1,0}}^T & \mathbf{X_{0,0}W_{0,1}}^T + X_{1,0}W_{1,1}^T \\
        X_{0,1}W_{0,0}^T + \mathbf{X_{1,1}W_{1,0}}^T & \mathbf{X_{0,1}W_{0,1}}^T + X_{1,1}W_{1,1}^T
    \end{bmatrix}
\end{align}
The bold-faced components represent the data that need to be communicated between the model-parallel instances.
For this operation, processes 0 and 1 can compute the partial sums $X_0W_{0,1}^T$ and $X_1W_{1,0}^T$, respectively, and communicate them to the neighboring process. Communication and computation can be overlapped. While transmitting these partial sums, the local terms $X_0W_{0,0}^T$ and $X_1W_{1,1}^T$ can be computed concurrently. The backward pass, which is just the transposed multiplication of the network weights with the Jacobi matrix, can be performed analogously.

\subsection{Four-way parallelism}
In 4-way parallelism, we partition the global data and parameter matrices along the last two dimensions, which in our case are the longitude and state variables. Hence, given four ranks, 0--3, each process holds data of dimensions [batch size, latitude, longitude/2, variables/2], represented as: $X = \begin{bmatrix}
    X_0 & X_1 \\ X_2 & X_3
\end{bmatrix}$. The weight matrices can be correspondingly represented  by $W = \begin{bmatrix}
    W_0 & W_1 \\ W_2 & W_3
\end{bmatrix}$.

A forward pass of a standard linear layer then yields:
\begin{align}
    XW^T &= \begin{bmatrix}
        X_0 & X_1 \\
        X_2 & X_3
    \end{bmatrix} 
    \begin{bmatrix}
        W_0^T & W_2^T \\
        W_1^T & W_3^T 
    \end{bmatrix} \\
    &= \begin{bmatrix}
        X_0 W_0^T + \mathbf{X_1 W_1^T} & \mathbf{X_0 W_2^T} + X_1 \mathbf{W_3^T} \\
        X_2 \mathbf{W_0^T} + \mathbf{X_3 W_1^T} & \mathbf{X_2 W_2^T} + X_3 W_3^T
    \end{bmatrix}
\end{align}
Pre-computation is performed where advantageous; for example, ranks 1 and 2 can compute $X_1W_1^T$ and $X_2W_2^T$ before transmitting the result to processes 0 and 3, respectively. Ranks 0 and 3 can compute the local terms $X_0W_0^T$ and $X_3W_3^T$ while waiting to receive the other partial sum, and so forth. While not implemented in this work, the model parallelism can be extended to arbitrary $n$-way parallelism by further splitting up the final dimensions into block-wise subdivisions.

\subsection{Data parallelism}
\label{subsec:dataparallel}
Conventional data parallelism can easily be combined with Jigsaw to train the model across multiple nodes. A reduction of the gradients needs to occur across the same portions of the model, i.e., given an $n$-way parallel model, all ranks \textit{r} with the same value of $r \% n$, where \% is the modulus operator, share the same parameters. 

\section{Implementation}
We utilize Jigsaw to parallelize all layers in WeatherMixer. Implementation is performed by overloading the PyTorch \texttt{autograd} function for a linear layer. Each permutation of $XW$, $XW^T$, $X^TW$ requires different communication patterns across the model-parallel processes in the forward--backward pass. We also eliminate a transpose operation in each mixing block by directly implementing a transposed MLP in which the forward pass is defined by $X^TW$ instead of $XW^T$. All communication is performed with MPI non-blocking point-to-point operations.
To take full advantage of the hierarchical network structure of most compute clusters, we perform intra-node tensor and domain parallelism, where fast, direct GPU-to-GPU communication is possible, and each GPU reads its own partition of the data independently. Inter-node parallelism occurs in the form of conventional data parallelism.
At its core, Jigsaw parallelism distributes matrix--matrix multiplication~\cite{vandeGeijn1995,Lee1997}. Hence, a fully model- and domain-parallel WM requires specialized implementations of convolutional layers, layer norms, and activation functions. The source code is open-source at \url{https://github.com/DeifiliaTo/Jigsaw_SC}. 

\paragraph{\textbf{Encoding and decoding layers:}}
The encoder and decoder were implemented via overloaded convolutional layers. A convolution is effectively a linear layer applied over a shifted window on, here, non-overlapping, spatially partitioned data.
As such, we implemented convolutions by first reshaping the input data into a series of non-overlapping patches and applying a linear layer over it. 

\paragraph{\textbf{Activation functions:}}
The GELU activation function is a point-wise operation, which is embarrassingly parallel; no synchronization is required. Utilization of Jigsaw on other activation functions such as Softmax would require an additional synchronization point. 

\paragraph{\textbf{Layer norms:}}
The layer norms in WeatherMixer are applied across each channel. In 2-way parallelism, PyTorch's native LayerNorm function can be used. However, in 4-way parallelism, the layer norm across ranks 0 and 2 will compute different gradients, despite learning normalization factors for the same channels. To ensure that the parameters of the layer norm stay synchronized as training progresses, a non-blocking pair-wise (beetween ranks 0 and 2; 1 and 3) reduce operation across the gradients in the layer norm is performed; this is similar to gradient-reduction in DP, but the reduction occurs across processes that otherwise do not contain the same parameters.



\paragraph{\textbf{Optimizer:}}
As each process holds its own shard of the model, the optimizers can update the parameters independently. No communication between the different model-parallel optimizers is required. The optimizer update step can be overlapped with loading the next training data sample onto the GPUs.

\paragraph{\textbf{Data loading:}}
Each model-parallel instance must load from the same data sample(s) at each training step. 
To ensure this, we set the same random seed for all model-parallel instances in the data loader (i.e., data-parallel instances use different seeds). 
The implementation of the data loader allows for each process to only read its relevant partition, including possible halos to resolve boundary conditions, allowing full parallel access of the data. The data loader applies zero-padding where necessary to ensure that the parameter dimensions remain constant across all model partitions.

\paragraph{\textbf{Data parallelism:}}
The PyTorch DDP wrapper takes care of the gradient reduction in the backward pass according to the description in \cref{subsec:dataparallel}, as well as parameter initialization. 

\section{Experimental Evaluation}
We conduct extensive experimental evaluation of our developed approaches by training WeatherMixer with and without Jigsaw parallelism in a number of different scenarios, evaluating both model predictive performance and scaling behaviour.
The ERA5 dataset obtained from WeatherBench2~\cite{Rasp2023} was used for training. In line with other atmospheric models, we use 6h-subsampled data at a global resolution of 0.25\degree{} and the following variables: surface variables are 10m u-velocity, 10 m v-velocity, 2m temperature, and mean sea level pressure; pressure-level variables are geopotential, specific humidity, temperature, and u- and v- velocity, at pressure levels of 1000, 925, 850, 700, 600, 500, 400, 300, 250, 200, 150, 100, and 50 \si{\hecto\pascal}. Constant inputs of soil type, topography, and land masks were also included as input.
Data from 1979--2017 were used for training and data from 2018 were used as the validation dataset. The weights for the different variables were adapted from Bi et al.~\cite{Bi_2023}. After receiving meteorological-grounded feedback that pressure levels close to the ground have higher importance, an additional pressure-level weighting was adapted: from high to low pressure levels, $[1, 1, 1, 1, 1, 1, 0.9, 0.8, 0.7, 0.6, 0.5, 0.4, 0.3]$.
The data were normalized with per-variable Z-score normalization. 
Models were evaluated based on latitude-weighted root mean squared error (RMSE), as is common practice~\cite{Rasp2023}.  
The Adam optimizer was used and models were trained for 100 epochs.  The initial learning rate for all parameters was $10^{-4}$, with the exception of the encoder and decoder layers, which are set to $2\cdot10^{-5}$ for improved stability. A ramped linear warm-up from $10^{-6}$ to the initial learning rate was used for the first epoch.
The learning rates were decayed to $10^{-5}$ with a cosine annealed learning-rate scheduler from epochs 2 to 100. Gradient clipping with a value of 1.0 was used to improve training stability.

To ensure stable dynamics over multi-step rollouts, we perform fine-tuning with a randomized technique. 
For each optimizer update step, a random rollout length $r$ is generated. 
In every training step, the ``processor'' (MLP-Mixer blocks) is carried out $r$ times, with each pass simulating one time step. Encoding and decoding are only performed once. This differs from a typical auto-regressive fine-tuning scheme \cite{Bi_2023,bodnar2024aurora,pathak2022} in which the encoder--decoder blocks are included in the rollout.


\subsection{Hardware and software specifications}
We ran all experiments on HoreKa, a distributed-memory, parallel hybrid supercomputer. 
Each compute node is equipped with two 38-core Intel Xeon Platinum 8368 processors at 2.4 GHz base and 3.4 GHz maximum turbo frequency, 512 GB local memory, two network adapters, and four NVIDIA A100-40 GPUs with 40 GB memory connected via NVLink. 
Inter-node communication uses a low-latency, non-blocking NVIDIA Mellanox InfiniBand 4X HDR interconnect with 200 GBit/s per port. 
HoreKa is equipped with Lenovo XClarity Controller sensors\footnote{\url{https://lenovopress.lenovo.com/lp1395-thinksystem-sd650-v2-server}} enabling whole-node, i.e., CPU, GPU, RAM, and Infiniband cards, energy usage measurements. The XClarity Controllers have a guaranteed accuracy of 97\% at a \qty{100}{\hertz} sampling rate.
All experiments used Python 3.11.2 with \texttt{CUDA}-enabled \texttt{PyTorch} 2.6.0~\citep{paszke2018pytorch}. The The NVIDIA Collective Communication Library (NCCL) backend was used for direct GPU-to-GPU communication. All models were trained with NVIDIA's TensorFloat-32 mixed precision. Certain scaling experiments were performed with uniform single precision.

\subsection{Model performance}
\label{subec:model-performance}
We evaluate model predictive performance for several scenarios, each highlighting a different aspect of model training.
\subsubsection{Equivalent usage}
In a realistic training scenario, users  have a fixed computational budget and dataset. With the added flexibility of model parallelism, WeatherMixer offers the option to shift between the number of data-parallel and model-parallel instances to be used during training. 
We demonstrate this by comparing model skill of three parallelization schemes of the same model, trained on the same fixed-sized dataset with  
\begin{enumerate*}
   \item a fixed compute budget of eight GPUs, and
   \item a fixed  model size and architecture, featuring 1 billion parameters, 3 MLP-Mixing blocks, and an embedding dimension ($d_\mathrm{emb}$) of 4320. In the MLP-Mixing blocks, a token mixing hidden dimension ($d_\mathrm{tok}$) of 8640 and a channel mixing hidden dimension ($d_\mathrm{ch}$) of 4320 is used.
\end{enumerate*} 
The 1-way model is a naive PyTorch implementation using 8 DP processes (global batch size of 8); the 2-way model employs Jigsaw parallelization and shards the model across two processes (global batch size of 4); and the 4-way model shards the model across four processes (global batch size of 2).

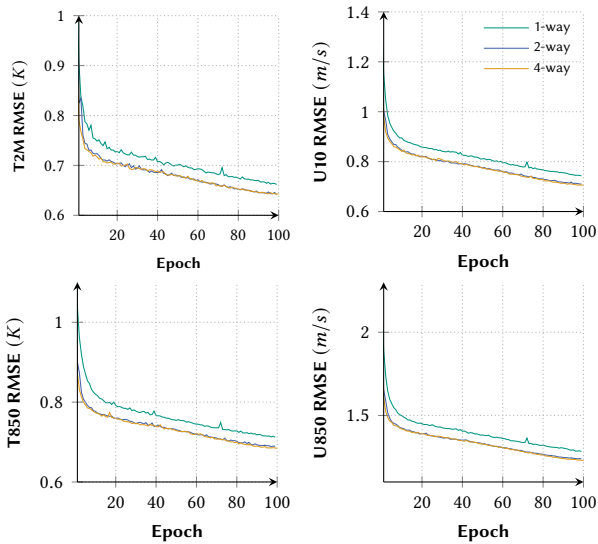
\begin{figure}
    \centering
    \begin{tikzpicture}
    \sffamily
    \begin{axis}[
        axis on top,
        width=0.5\linewidth,
        height=0.5\linewidth,
        xlabel=Epoch,
        ylabel=T2M RMSE $(K)$,
        label style={font=\scriptsize\bfseries},
        tick label style={
            font=\scriptsize\bfseries\sffamily,
            /pgf/number format/1000 sep=
        },
        axis x line=bottom,
        axis y line=left,
        xmode=linear,
        ymode=linear,
        xmin=1,
        xmax=100,
        ymin=0.6,
        ymax=1,
        grid=major,
        grid style={ 
            draw=kit-gray30,
            densely dotted
        }
    ]
        \addplot[
            color=kit-green, 
            point meta=explicit symbolic, 
            nodes near coords,
            nodes near coords align={
                yshift=6.3
            },
            every node near coord/.style={
                font=\tiny,
                color=black
            },
        ] table [x=epoch, y=t2m, col sep=comma] {figures/data/rmse_one_0.csv};
        \addplot[
            color=kit-blue, 
            point meta=explicit symbolic, 
            nodes near coords,
            nodes near coords align={
                yshift=6.3
            },
            every node near coord/.style={
                font=\tiny,
                color=black
            },
        ] table [x=epoch, y=t2m, col sep=comma] {figures/data/rmse_two_0.csv};
        \addplot[
            color=kit-orange, 
            point meta=explicit symbolic, 
            nodes near coords,
            nodes near coords align={
                yshift=6.3
            },
            every node near coord/.style={
                font=\tiny,
                color=black
            },
        ] table [x=epoch, y=t2m, col sep=comma] {figures/data/rmse_four_0.csv};
    \end{axis}
\end{tikzpicture}
\begin{tikzpicture}
    \sffamily
    \begin{axis}[
        axis on top,
        width=0.5\linewidth,
        height=0.5\linewidth,
        xlabel=Epoch,
        ylabel=U10 RMSE $(m/s)$,
        label style={font=\footnotesize\bfseries},
        tick label style={
            font=\footnotesize\bfseries\sffamily,
            /pgf/number format/1000 sep=
        },
        axis x line=bottom,
        axis y line=left,
        xmode=linear,
        ymode=linear,
        xmin=1,
        xmax=100,
        ymin=0.6,
        ymax=1.4,
        grid=major,
        grid style={ 
            draw=kit-gray30,
            densely dotted
        },
        legend style = {
            draw=none,
            fill=none,
            font=\footnotesize\sffamily,
            at={(1.0,1)},
            anchor=north east,
            nodes={scale=0.8, transform shape}
        }
    ]
        \addplot[
            color=kit-green, 
            point meta=explicit symbolic, 
            nodes near coords,
            nodes near coords align={
                yshift=6.3
            },
            every node near coord/.style={
                font=\tiny,
                color=black
            },
        ] table [x=epoch, y=u10, col sep=comma] {figures/data/rmse_one_0.csv};
        \addlegendentry{1-way}
        \addplot[
            color=kit-blue, 
            point meta=explicit symbolic, 
            nodes near coords,
            nodes near coords align={
                yshift=6.3
            },
            every node near coord/.style={
                font=\tiny,
                color=black
            },
        ] table [x=epoch, y=u10, col sep=comma] {figures/data/rmse_two_0.csv};
        \addlegendentry{2-way}
        \addplot[
            color=kit-orange, 
            point meta=explicit symbolic, 
            nodes near coords,
            nodes near coords align={
                yshift=6.3
            },
            every node near coord/.style={
                font=\tiny,
                color=black
            },
        ] table [x=epoch, y=u10, col sep=comma] {figures/data/rmse_four_0.csv};
        \addlegendentry{4-way}
    \end{axis}
\end{tikzpicture}

\begin{tikzpicture}
    \sffamily
    \begin{axis}[
        axis on top,
        width=0.5\linewidth,
        height=0.5\linewidth,
        xlabel=Epoch,
        ylabel=T850 RMSE $(K)$,
        label style={font=\footnotesize\bfseries},
        tick label style={
            font=\footnotesize\bfseries\sffamily,
            /pgf/number format/1000 sep=
        },
        axis x line=bottom,
        axis y line=left,
        xmode=linear,
        ymode=linear,
        xmin=1,
        xmax=100,
        ymin=0.6,
        ymax=1.1,
        grid=major,
        grid style={ 
            draw=kit-gray30,
            densely dotted
        }
    ]
        \addplot[
            color=kit-green, 
            point meta=explicit symbolic, 
            nodes near coords,
            nodes near coords align={
                yshift=6.3
            },
            every node near coord/.style={
                font=\tiny,
                color=black
            },
        ] table [x=epoch, y=t850, col sep=comma] {figures/data/rmse_one_0.csv};
        \addplot[
            color=kit-blue, 
            point meta=explicit symbolic, 
            nodes near coords,
            nodes near coords align={
                yshift=6.3
            },
            every node near coord/.style={
                font=\tiny,
                color=black
            },
        ] table [x=epoch, y=t850, col sep=comma] {figures/data/rmse_two_0.csv};
        \addplot[
            color=kit-orange, 
            point meta=explicit symbolic, 
            nodes near coords,
            nodes near coords align={
                yshift=6.3
            },
            every node near coord/.style={
                font=\tiny,
                color=black
            },
        ] table [x=epoch, y=t850, col sep=comma] {figures/data/rmse_four_0.csv};
    \end{axis}
\end{tikzpicture}
\begin{tikzpicture}
    \sffamily
    \begin{axis}[
        axis on top,
        width=0.5\linewidth,
        height=0.5\linewidth,
        xlabel=Epoch,
        ylabel=U850 RMSE $(m/s)$,
        label style={font=\footnotesize\bfseries},
        tick label style={
            font=\footnotesize\bfseries\sffamily,
            /pgf/number format/1000 sep=
        },
        axis x line=bottom,
        axis y line=left,
        xmode=linear,
        ymode=linear,
        xmin=1,
        xmax=100,
        ymin=1.1,
        ymax=2.3,
        grid=major,
        grid style={ 
            draw=kit-gray30,
            densely dotted
        }
    ]
        \addplot[
            color=kit-green, 
            point meta=explicit symbolic, 
            nodes near coords,
            nodes near coords align={
                yshift=6.3
            },
            every node near coord/.style={
                font=\tiny,
                color=black
            },
        ] table [x=epoch, y=u850, col sep=comma] {figures/data/rmse_one_0.csv};
        \addplot[
            color=kit-blue, 
            point meta=explicit symbolic, 
            nodes near coords,
            nodes near coords align={
                yshift=6.3
            },
            every node near coord/.style={
                font=\tiny,
                color=black
            },
        ] table [x=epoch, y=u850, col sep=comma] {figures/data/rmse_two_1.csv};
        \addplot[
            color=kit-orange, 
            point meta=explicit symbolic, 
            nodes near coords,
            nodes near coords align={
                yshift=6.3
            },
            every node near coord/.style={
                font=\tiny,
                color=black
            },
        ] table [x=epoch, y=u850, col sep=comma] {figures/data/rmse_four_1.csv};
    \end{axis}
\end{tikzpicture}
    \caption{Validation RMSE values for key variables throughout training for 1-billion-parameter models with equivalent architectures across 1-, 2- and 4-way parallel models. Values reported are the RMSE for a 6-hour forecast.}
    \label{fig:rmse_comparison}
    \Description{This is a description}
\end{figure}
Figure~\ref{fig:rmse_comparison} shows the validation RMSE for key variables during training. For all key variables, the parallel models show better predictive performance than the naive model. Training time on eight A100-40 GB GPUs amounts to 104, 113, and 155 minutes/epoch, for the 1-, 2-, 4-way parallel models. While parallelizing the 4-way model incurs substantial increase in computational time, the 2-way model achieves RMSE values that are lower by 2-9\%, by only increasing the required computational resources by 9\%. 
As the architecture in the MLP-Mixer blocks are identical between the naive and the parallel implementations, we hypothesize that these results are due to large-batch effects~\cite{keskar2016large}. Previous work has reported that AI-based atmospheric models seem to exhibit large batch effects very quickly, i.e., training runs that are performed with small global batch sizes and more optimizer update steps will converge to lower loss values than models with large global batch sizes~\cite{To2024}. 
As such, Figure~\ref{fig:rmse_comparison} shows empirical results that training parallel models improves convergence due to a higher number of optimizer steps, mitigating the problematic large batch effect.

\subsubsection{One-step model performance}
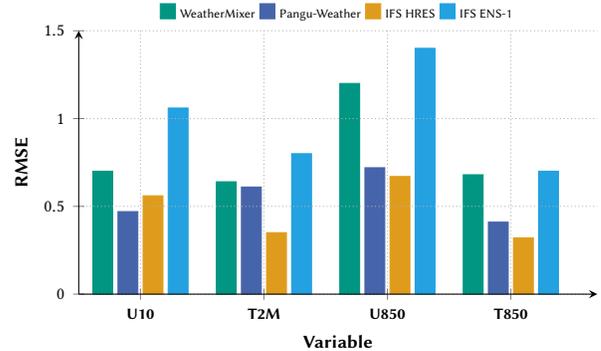
\begin{figure}
    \centering
    \begin{tikzpicture}
    \sffamily
    \begin{axis}[
        axis on top,
        ybar,
        bar width=7.5pt,
        width=1\linewidth,
        height=0.6\linewidth,
        xlabel=Variable,
        ylabel=RMSE,
        label style={font=\footnotesize\bfseries},
        tick label style={
            font=\scriptsize\bfseries\sffamily,
            /pgf/number format/1000 sep=
        },
        axis x line=bottom,
        axis y line=left,
        xtick={1,2,3,4},
        xticklabels={U10, T2M, U850, T850},
        ymode=linear,
        xmin=0.5,
        xmax=4.7,
        ymin=0,
        ymax=1.5,
        grid=major,
        grid style={ 
            draw=kit-gray30,
            densely dotted
        },
        legend columns=4,
        legend cell align=left,
        legend pos=north east,
        legend style = {
            draw=none,
            fill=none,
            font=\scriptsize\sffamily,
            at={(0.5, 1.0)},
            anchor=south,
            nodes={scale=0.8, transform shape}
        },
         legend image code/.code={
            \draw[#1] (0cm,-0.05cm) rectangle (0.2cm, 0.15cm);
        }
    ]
        \addplot[
            color=kit-green, 
            fill=kit-green,
            every node near coord/.style={
                font=\tiny,
                color=black
            },
        ] coordinates{
            (1,0.7) 
            (2,0.64)
            (3,1.2)
            (4,0.68)
        };
        \addplot[
            color=kit-blue,
            fill=kit-blue,
            every node near coord/.style={
                font=\tiny,
                color=black
            },
        ] coordinates{
            (1,0.47)
            (2,0.61)
            (3,0.72)
            (4,0.41)
        };
        \addplot[
            color=kit-orange, 
            fill=kit-orange,
            every node near coord/.style={
                font=\tiny,
                color=black
            },
        ] coordinates{
            (1,0.56)
            (2,0.35)
            (3,0.67)
            (4,0.32)
        };
        \addplot[
            color=kit-cyan, 
            fill=kit-cyan,
            every node near coord/.style={
                font=\tiny,
                color=black
            },
        ] coordinates{
            (1,1.06)
            (2,0.8)
            (3,1.4)
            (4,0.7)
        };
        \legend{WeatherMixer,Pangu-Weather,IFS HRES, IFS ENS-1}
    \end{axis}
\end{tikzpicture}
    \caption{Validation RMSE of best-performing WM model trained with 2-way MP across 8 GPUs. Reference benchmarks are Pangu-Weather, IFS HRES, and IFS ENS---1 ensemble member reported from WeatherBench~\cite{Rasp2023}.}
    \label{fig:onestep}
    \Description{This is a description}
\end{figure}

Figure~\ref{fig:onestep} shows the validation RMSE values for key variables of the best-performing 1-billion-parameter WM model trained with 2-way model parallelism across 8 GPUS (four data-parallel instances). For all variables, we achieve a forecast skill higher than IFS--1 ensemble member. Since the WM models have only been trained on a 6h-subset of the dataset, i.e., 16\% of the available data, this represents an upper-bound of the RMSE values achievable by this model architecture as per neural scaling laws~\cite{kaplan2020scaling}.

\subsubsection{Rolled-out model performance}
\begin{figure}
    \centering
    \begin{tikzpicture}
    \sffamily
    \begin{axis}[
        axis on top,
        width=0.5\linewidth,
        height=0.6\linewidth,
        xlabel=Lead time $(h)$,
        ylabel=U10 RMSE $(m/s)$,
        label style={font=\footnotesize\bfseries},
        tick label style={
            font=\footnotesize\bfseries\sffamily,
            /pgf/number format/1000 sep=
        },
        axis x line=bottom,
        axis y line=left,
        xmode=linear,
        ymode=linear,
        xmin=6,
        xmax=120,
        ymin=0,
        ymax=3.5,
        grid=major,
        grid style={ 
            draw=kit-gray30,
            densely dotted
        },
        legend style={
            draw=none,
            fill=none,
            font=\footnotesize\sffamily,
            at={(1,1)},
            anchor=north east,
            nodes={scale=0.8, transform shape}
        }
    ]
        \addplot[
            color=kit-green, 
            mark=*, 
            mark options={
                scale=0.5, 
            },
            point meta=explicit symbolic, 
            nodes near coords,
            nodes near coords align={
                yshift=6.3
            },
            every node near coord/.style={
                font=\tiny,
                color=black
            },
        ] table [x=rollout_time, y=u10, col sep=comma] {figures/data/wm_rollout.csv};
        \addplot[
            color=kit-blue, 
            mark=*, 
            mark options={
                scale=0.5, 
            },
            point meta=explicit symbolic, 
            nodes near coords,
            nodes near coords align={
                yshift=6.3
            },
            every node near coord/.style={
                font=\tiny,
                color=black
            },
        ] table [x=lead_time, y=u10_pangu_era5, col sep=comma] {figures/data/reference_scores.csv};
        \addplot[
            color=kit-orange, 
            mark=*, 
            mark options={
                scale=0.5, 
            },
            point meta=explicit symbolic, 
            nodes near coords,
            nodes near coords align={
                yshift=6.3
            },
            every node near coord/.style={
                font=\tiny,
                color=black
            },
        ] table [x=lead_time, y=u10_ifs_analysis, col sep=comma] {figures/data/reference_scores.csv};
        \addplot[
            color=kit-cyan, 
            mark=*, 
            mark options={
                scale=0.5, 
            },
            point meta=explicit symbolic, 
            nodes near coords,
            nodes near coords align={
                yshift=6.3
            },
            every node near coord/.style={
                font=\tiny,
                color=black
            },
        ] table [x=lead_time, y=u10_ens1, col sep=comma] {figures/data/reference_scores.csv};
    \end{axis}
\end{tikzpicture}
\begin{tikzpicture}
    \sffamily
    \begin{axis}[
        axis on top,
        width=0.5\linewidth,
        height=0.6\linewidth,
        xlabel=Lead time $(h)$,
        ylabel=U850 RMSE $(m/s)$,
        label style={font=\footnotesize\bfseries},
        tick label style={
            font=\footnotesize\bfseries\sffamily,
            /pgf/number format/1000 sep=
        },
        axis x line=bottom,
        axis y line=left,
        xmode=linear,
        ymode=linear,
        xmin=6,
        xmax=120,
        ymin=0,
        ymax=4.9,
        grid=major,
        grid style={ 
            draw=kit-gray30,
            densely dotted
        },
        legend style = {
            draw=none,
            fill=none,
            font=\scriptsize\sffamily,
            at={(1, 0)},
            anchor=south east,
            nodes={scale=0.8, transform shape}
        },
        legend cell align={left}
    ]
        \addplot[
            color=kit-green, 
            mark=*, 
            mark options={
                scale=0.5, 
            },
            point meta=explicit symbolic, 
            nodes near coords,
            nodes near coords align={
                yshift=6.3
            },
            every node near coord/.style={
                font=\tiny,
                color=black
            },
        ] table [x=rollout_time, y=u850, col sep=comma] {figures/data/wm_rollout.csv};
        \addlegendentry{WeatherMixer}
        \addplot[
            color=kit-blue, 
            mark=*, 
            mark options={
                scale=0.5, 
            },
            point meta=explicit symbolic, 
            nodes near coords,
            nodes near coords align={
                yshift=6.3
            },
            every node near coord/.style={
                font=\tiny,
                color=black
            },
        ] table [x=lead_time, y=u850_pangu_era5, col sep=comma] {figures/data/reference_scores.csv};
        \addlegendentry{Pangu-Weather}
        \addplot[
            color=kit-orange, 
            mark=*, 
            mark options={
                scale=0.5, 
            },
            point meta=explicit symbolic, 
            nodes near coords,
            nodes near coords align={
                yshift=6.3
            },
            every node near coord/.style={
                font=\tiny,
                color=black
            },
        ] table [x=lead_time, y=u850_ifs_analysis, col sep=comma] {figures/data/reference_scores.csv};
        \addlegendentry{IFS HRES}
        \addplot[
            color=kit-cyan, 
            mark=*, 
            mark options={
                scale=0.5, 
            },
            point meta=explicit symbolic, 
            nodes near coords,
            nodes near coords align={
                yshift=6.3
            },
            every node near coord/.style={
                font=\tiny,
                color=black
            },
        ] table [x=lead_time, y=u850_ens1, col sep=comma] {figures/data/reference_scores.csv};
        \addlegendentry{IFS ENS-1}
    \end{axis}
\end{tikzpicture}
    \caption{Validation RMSE of best-performing  WM model trained with 2-way across 8 GPUs, fine-tuned over long rollouts~\cite{Rasp2023}.}
    \label{fig:rollout}
    \Description{This is a description}
\end{figure}
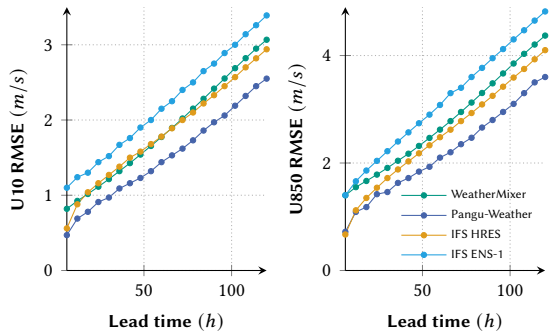

To test the performance over longer time scales, the 2-way MP, 1-billion-parameter model was rolled out over twenty 6h-steps, up a lead time of 120 h (cf. Figure~\ref{fig:rollout}). We emphasize that fine-tuning of a longer rollout is only made possible with MP, since the memory requirements exceed the capacity of a single accelerator.

\subsection{Scaling}
Scaling experiments were conducted by linearly increasing the amount of work, defined as floating point operations (FLOPs) per forward pass of the model per GPU. The models were defined by changing the embedding dimension of the encoder and the hidden dimension in the MLPs for the token mixing ($d_\mathrm{tok}$). The hidden dimension of the channel mixing MLPs was equal to the embedding dimension $d_\mathrm{ch} = d_\mathrm{emb}$. The number of parameters was roughly increased linearly with the increasing computational workload. The number of layers, i.e., matrix--matrix multiplications, was kept constant throughout all models. Details of the model architectures are shown in Table~\ref{tab:model_size}.
The workload was increased from 0.25 TFLOPs/forward pass/GPU up to 16 TFLOPs/forward pass/GPU. The upper limit was chosen to correspond to the maximum model size that would fit in the memory of a single GPU, which is roughly 1.4 billion parameters. This was calculated by considering the dimensions of the matrix--matrix operations within the model. Due to the simple architecture, the backward pass was considered to have two times the number of FLOPs as the forward pass.
FLOPs incurred in layer norms, reductions, and dropout layers were assumed to be negligible compared to the model size.
Experiments were performed with both uniform single precision and TensorFloat-32 (TF32) mixed precision; these precisions have a reported peak performance of 19.5 TFLOPs/s/GPU and 156 TFLOPs/s/GPU, respectively, on the NVIDIA A100-40GB accelerators.


\begin{table}[h!]
  \caption{Details of model architectures in scaling experiments.}
  \label{tab:model_size}
  \begin{tabular}{crrrrr}
    \toprule
    \textbf{Model} \# & \textbf{TFLOPs} & \textbf{Params (mil)} & $\boldsymbol{d_\mathrm{emb}}$ & $\boldsymbol{d_\mathrm{tok}}$ & $\boldsymbol{d_\mathrm{ch}}$ \\
    \midrule
    1 & 0.25 & 60 & 240 & 540 & 240 \\
    2 & 0.5  & 230 & 512 & 2,160 & 512 \\
    3 & 1 & 240 & 896 & 2,160 & 896 \\
    4 & 2 & 260 & 1,600 & 2,160 & 1,600 \\
    5 & 4 & 500 & 2,192 & 4,320 & 2,192 \\
    6 & 8 & 980 & 2,832 & 8,640 & 2,832 \\
    7 & 16 & 1,400 & 4,896 & 8,640 & 4,896 \\
    8 & 32 & 2,000 & 6,064 & 17,280 & 6,064 \\
    9 & 64 & 2,600 & 10,352 & 17,280 & 10,352 \\  
  \bottomrule
\end{tabular}
\end{table}

Performance was assessed by measuring the average time required for 10 epochs. Each epoch comprises, for all data samples in the data subset, 
\begin{enumerate*}
\item CPUs pre-fetching data from disk storage and normalization of the data,
\item loading the data from CPU to GPU,
\item forward pass of the model,
\item backward pass of the model,
\item reduction of gradients if required,
\item and optimizer update step.
\end{enumerate*}
A subset of 500 data samples were included for the uniform precision experiments, and 1,250 for the mixed precision experiments.

\subsubsection{Roofline analysis}
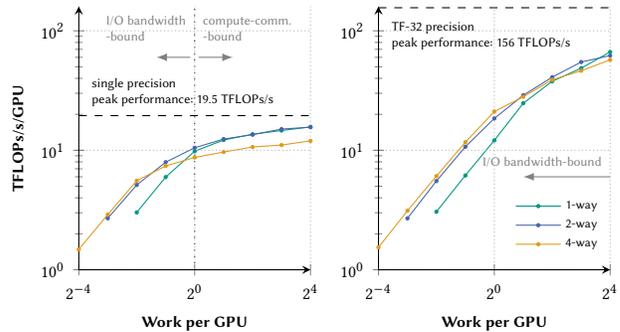
\begin{figure}
    \centering
    \begin{tikzpicture}
    \sffamily
    \begin{axis}[
        axis on top,
        width=0.55\linewidth,
        height=0.6\linewidth,
        xlabel=Work per GPU,
        ylabel=TFLOPs/s/GPU,
        label style={font=\scriptsize\bfseries},
        tick label style={
            font=\scriptsize\bfseries\sffamily,
            /pgf/number format/1000 sep=
        },
        axis x line=bottom,
        axis y line=left,
        xmode=log,
        ymode=log,
        log basis x=2,
        xmin=0.0625,
        xmax=16,
        ymin=1,
        ymax=160,
        scaled y ticks=false,
        grid=major,
        grid style={ 
            draw=kit-gray30,
            densely dotted
        }
    ]
        \addplot[
            color=kit-green, 
            mark=*, 
            mark options={
                scale=0.3, 
            },
            point meta=explicit symbolic, 
            nodes near coords,
            nodes near coords align={
                yshift=6.3
            },
            every node near coord/.style={
                font=\tiny,
                color=black
            },
        ] table [x=Work_per_GPU, y=one_TFLOPss, col sep=comma] {figures/data/roofline_uniform.csv};
        \addplot[
            color=kit-blue, 
            mark=*, 
            mark options={
                scale=0.3, 
            },
            point meta=explicit symbolic, 
            nodes near coords,
            nodes near coords align={
                yshift=6.3
            },
            every node near coord/.style={
                font=\tiny,
                color=black
            },
        ] table [x=Work_per_GPU, y=two_TFLOPss, col sep=comma] {figures/data/roofline_uniform.csv};
        \addplot[
            color=kit-orange, 
            mark=*, 
            mark options={
                scale=0.3, 
            },
            point meta=explicit symbolic, 
            nodes near coords,
            nodes near coords align={
                yshift=6.3
            },
            every node near coord/.style={
                font=\tiny,
                color=black
            },
        ] table [x=Work_per_GPU, y=four_TFLOPss, col sep=comma] {figures/data/roofline_uniform.csv};
        \addplot[
        color=black,
        domain=0.0625:16,
        dashed
        ] {19.5};
        \node[anchor=south west, font=\tiny, color=kit-gray, align=left] at (axis cs:0.07, 20) {single precision\\peak performance: 19.5 TFLOPs/s};

        \addplot[dotted, samples=100, smooth,domain=0:6,black] coordinates {(1,1)(1,160)};
        \node[anchor=south east,kit-gray50, align=left, font=\tiny] (source1) at (1, 70){I/O bandwidth\\-bound};
        \node[anchor=south west,kit-gray50, align=left, font=\tiny] (source2) at (1, 70){compute-comm.\\-bound};
        \draw[-latex,kit-gray50](0.9,60)--(0.4,60);
        \draw[-latex,kit-gray50](1.1,60)--(2.5,60);
    \end{axis}
\end{tikzpicture}
\begin{tikzpicture}
    \sffamily
    \begin{axis}[
        axis on top,
        width=0.55\linewidth,
        height=0.6\linewidth,
        xlabel=Work per GPU,
        label style={font=\scriptsize\bfseries},
        tick label style={
            font=\scriptsize\bfseries\sffamily,
            /pgf/number format/1000 sep=
        },
        axis x line=bottom,
        axis y line=left,
        xmode=log,
        ymode=log,
        log basis x=2,
        xmin=0.0625,
        xmax=16,
        ymin=1,
        ymax=160,
        grid=major,
        grid style={ 
            draw=kit-gray30,
            densely dotted
        },
        legend style = {
            draw=none,
            fill=none,
            font=\scriptsize\sffamily,
            at={(1.0,0.3)},
            anchor=north east,
            nodes={scale=0.8, transform shape}
        }
    ]
        \addplot[
            color=kit-green, 
            mark=*, 
            mark options={
                scale=0.3, 
            },
            point meta=explicit symbolic, 
            nodes near coords,
            nodes near coords align={
                yshift=6.3
            },
            every node near coord/.style={
                font=\tiny,
                color=black
            },
        ] table [x=Work_per_GPU, y=one_TFLOPss, col sep=comma] {figures/data/roofline_mixed.csv};
        \addlegendentry{1-way}
        \addplot[
            color=kit-blue, 
            mark=*, 
            mark options={
                scale=0.3, 
            },
            point meta=explicit symbolic, 
            nodes near coords,
            nodes near coords align={
                yshift=6.3
            },
            every node near coord/.style={
                font=\tiny,
                color=black
            },
        ] table [x=Work_per_GPU, y=two_TFLOPss, col sep=comma] {figures/data/roofline_mixed.csv};
        \addlegendentry{2-way}
        \addplot[
            color=kit-orange, 
            mark=*, 
            mark options={
                scale=0.3, 
            },
            point meta=explicit symbolic, 
            nodes near coords,
            nodes near coords align={
                yshift=6.3
            },
            every node near coord/.style={
                font=\tiny,
                color=black
            },
        ] table [x=Work_per_GPU, y=four_TFLOPss, col sep=comma] {figures/data/roofline_mixed.csv};
        \addlegendentry{4-way}
        \addplot[
        color=black,
        domain=0.0625:16,
        dashed
        ] {156};
        \node[anchor=north west, font=\tiny, color=kit-gray, align=left] at (axis cs:0.07, 140) {TF-32 precision \\ peak performance: 156 TFLOPs/s};
        \node[anchor=east,kit-gray50] (source1) at (16, 8){\tiny I/O bandwidth-bound};
        \draw[-latex,kit-gray50](16,6)--(2,6);
    \end{axis}
\end{tikzpicture}
    \caption{Roofline plot for training models with 1-, 2-, 4-way parallelism. Performance for computations (left) in uniform single precision; (right) in mixed TF32 precision.}
    \label{fig:roofline}
    \Description{This is a description}
\end{figure}
The roofline plot in Figure~\ref{fig:roofline} shows two regimes: one that is bound by data access from storage (I/O-bandwidth limited); and one that is bound by the computational load of the forward--backward pass (computation--communication-limited), which increases with model size. Figure~\ref{fig:roofline} (left) shows the performance for uniform single precision and Figure~\ref{fig:roofline} (right) shows the performance of TF32 precision.
As a baseline without MP (1-way), we reach 81\% of theoretical peak performance with uniform single precision, and 43\% with mixed precision. These are strong baselines for further analyses, as these measurements represent performance within an operational machine-learning training run.

For uniform single precision, (Figure~\ref{fig:roofline} (left)) shows that training is I/O-bandwidth-limited when the model is small. The computation--communication-limited regime is reached at a modest model size of 1 TFLOP per forward pass, corresponding to 240 million parameters. In this regime, the 4-way parallel model obtains a lower performance than the non-MP (1-way) baseline, as communication costs dominate. However, we see remarkable performance of the 2-way model, which reaches near-unity computational performance (80\% theoretical peak performance) when compared with the 1-way baseline, indicating that communication is efficiently overlapped with computation.

The roofline plots for the mixed-precision computations (Figure~\ref{fig:roofline} (right) show that the performance is I/O-bandwidth-limited for all model sizes, i.e., GPUs run out-of-memory before reaching the computation--communication-limited regime. Accordingly, we do not expect any performance gains from even lower precision computations. 
Models within the I/O-bandwidth-limited regime benefit greatly from Jigsaw's domain parallelism: By reducing the volume of data that needs to be loaded, we reduce the size of the bottleneck.
As such, for small models, arithmetic throughput is higher for parallel models than the non-MP implementation; and for the highest computational load, we observe 43\%, 40\%, 37\% of peak performance for the 1-, 2-, 4-way models, respectively, meaning that the 4-way parallel model has a relative performance of 86\% compared to the non-MP baseline.

\subsubsection{Strong scaling}
\begin{figure}
    \centering
    \input{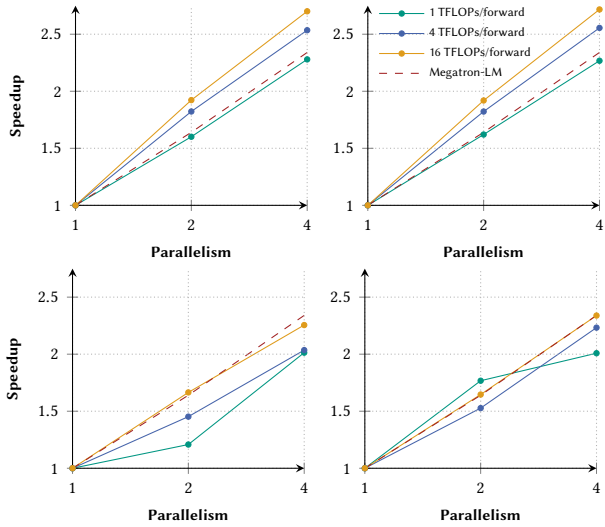}
    \caption{Strong scaling experiments for training 240 million, 500 million, 1.4 billion parameter models, corresponding to 1, 4, 16 TFLOPs/forward pass. (top left) No data loading, uniform single precision; (top right) Full training loop, uniform single precision; (bottom left) No data loading, mixed TF32 precision; (bottom right) Full training loop, mixed TF32 precision.}
    \label{fig:strong_scaling}
    \Description{This needs a description}
\end{figure}
To isolate the performance of Jigsaw parallelism from the data loading effect and to highlight Jigsaw's domain-parallel advantages, we present scaling results in two modes: one normal training loop, and one in which data loading from CPUs to GPUs is completely excluded. These represent the scaling results if data loading is not a bottleneck, i.e., in training cases where the input data are small compared to the model size.
This was realized by loading only one data sample at the first iteration of the training loop, and performing subsequent forward and backward passes on this sample.
For strong scaling experiments, the local and global batch size were kept constant at one, and models were distributed among 1, 2, and 4 processes. The size of the dataset was kept constant across all experiments. Results are shown on increasing model sizes: 1, 4, 16 TFLOPs/forward pass, corresponding to 240, 500, 1\,400 million parameters in total (models 3, 5, 7 in Table~\ref{tab:model_size}).

For uniform single precision strong scaling, shown in Figure~\ref{fig:strong_scaling} (top right) and corresponding with the roofline plot in Figure~\ref{fig:roofline} (left), we find that even the smallest model with 1 TFLOP/forward pass reaches the computation--communication-limited regime. This allows for communication synchronization to occur simultaneously with computation, resulting in exceptionally good scaling: for a 1.4-billion-parameter model, speedup with Jigsaw is 1.9 and 2.7 for 2- and 4-way parallel models, surpassing Megatron-LM's results of 1.6 and 2.3, respectively, on their 1.2 billion parameter model~\cite{shoeybi2020}.

The mixed-precision TF32 results in Figure~\ref{fig:strong_scaling} (bottom right) show scaling results for models that are I/O-bandwidth-limited, whereas Figure~\ref{fig:strong_scaling} (bottom left) shows scaling of computation-communication-limited models, i.e., data loading is neglected. Scaling efficiency increases as the workload increases, but for the largest 1.4-billion-parameter model, strong scaling results are slightly worse than Megatron-LM. This is expected, since Jigsaw requires all MP instances to be synchronized at every matrix--matrix multiplication, providing a significant bottleneck. 
In practice, however, operational model training is I/O-bandwidth-limited (Figure~\ref{fig:strong_scaling} (bottom right)). In this regime, the advantages of Jigsaw's domain parallelism and partitioned data loading dominate, yielding strong scaling competitive to Megatron-LM. 
The ``kink'' in the smallest model can be explained by studying the roofline plot in Figure~\ref{fig:roofline} (right). Here, the model is small and we are completely in the I/O-bandwidth limited regime, and hence speedup is bounded by the speed of data loading. When increasing model size, the computation becomes the dominating factor again and we observe higher speedup.


\subsubsection{Weak scaling}
\begin{figure}
    \centering
    \input{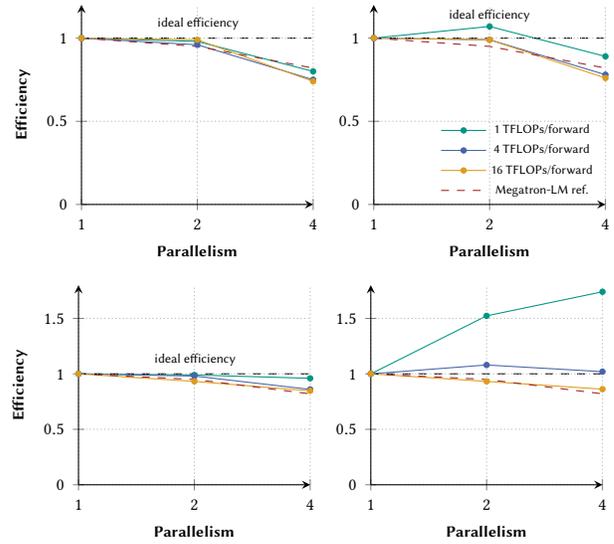}
    \caption{Weak scaling experiments for training  models with to 1, 4, 16 TFLOPs/GPU/forward pass. (top left) No data loading, uniform single precision; (top right) Full training loop, uniform single precision; (bottom left) No data loading, mixed TF32 precision; (bottom right) Full training loop, mixed TF32 precision.}
    \label{fig:weak_scaling}
    \Description{This needs a description}
\end{figure}

In weak scaling experiments, local and global batch size were kept constant at 1. The model size was increased by keeping the number of FLOPs/GPU constant.
Results are shown on three sets of models: 1, 4, 16 TFLOPs/ forward pass/GPU, corresponding to 240, 500, 1\,400 million global parameters in the baseline. The dataset size is kept constant across all experiments. 
In operational training, Figure~\ref{fig:weak_scaling} (top right, bottom right), we see superscalar results in models that are strictly I/O-bandwidth-limited, as parallel models benefit from partitioned data loading. 
Figure~\ref{fig:weak_scaling} (bottom right) shows that, in small models (1 TFLOP/forward pass/GPU) that are purely I/O-bandwidth limited, higher degrees of parallelism yield superscalar efficiencies in weak scaling. In larger models (4 TFLOPs/forward pass/GPU), superscaling efficiency is still possible, but computational costs start to dominate the 4-way parallel model.
In the largest model (64 TFLOPs/forward pass), communication overhead dominates and superscaling is no longer observed. We show weak scaling efficiency of 86\% for the 4-way parallel model, surpassing Megatron-LM's baseline of 82\% efficiency.

\subsubsection{Data-Parallel Training}
System-wide scaling experiences were performed to study the effects of data-parallel training on a large scale. WeatherMixer was trained on up to 256 GPUs.
For up to four GPUs, weak model scaling was performed: one model instance was trained on one, two, and four GPUs. The workload was held constant at 16 TFLOPs/forward pass/GPU, and the model size was increased up to 4 GPUs. Weak inter-node data scaling was performed by increasing the size of the dataset. The baselines for each scaling experiment is the corresponding model-parallel group, with a global batch size of 1.
In the base cases (1, 2, 4 GPUs), 100 data samples were included in the dataset per epoch.  Details on the computational workload for each case, the number of parameters, and the number of data-parallel model instances when scaling to multiple nodes are presented in Table~\ref{tab:ddp}. 

\begin{table}[h!]
  \caption{Number of data-parallel model instances}
  \label{tab:ddp}
  \resizebox{\linewidth}{!}{%
  \begin{tabular}{lccccccccc}
    \toprule
    & \textbf{TFLOPs} & \textbf{Params (mil)} & \multicolumn{7}{c}{\textbf{GPUs}} \\
    \cmidrule(lr){4-10}
    & & & 1 & 2 & 4 & 8 & 16 & ... & 256 \\
    \midrule
    \textbf{1-way} & 16 & 1000 & 1 & 2 & 4 & 8 & 16 & ... & 256 \\
    \textbf{2-way} & 32 & 1400 & - & 1 & 2 & 4 & 8 & ... & 128 \\
    \textbf{4-way} & 64 & 2400 & - & - & 1 & 2 & 4 & ... & 64 \\
  \bottomrule
\end{tabular}}
\end{table}

\begin{figure}
    \centering
    \begin{tikzpicture}
    \sffamily
    \begin{axis}[
        axis on top,
        width=0.9\linewidth,
        height=0.6\linewidth,
        xlabel=GPUs,
        ylabel=Efficiency,
        label style={font=\footnotesize\bfseries},
        tick label style={
            font=\footnotesize\bfseries\sffamily,
            /pgf/number format/1000 sep=
        },
        axis x line=bottom,
        axis y line=left,
        xmode=log,
        ymode=linear,
        log basis x=2,
        log ticks with fixed point,
        xmin=8,
        xmax=256,
        ymin=0,
        ymax=1,
        grid=major,
        grid style={ 
            draw=kit-gray30,
            densely dotted
        },
        legend style = {
            draw=none,
            fill=none,
            font=\footnotesize\sffamily,
            at={(1.0,0.4)},
            anchor=north east,
            nodes={scale=0.8, transform shape}
        }
    ]
        \addplot[
            color=kit-green, 
            mark=*, 
            mark options={
                scale=0.5, 
            },
            point meta=explicit symbolic, 
            nodes near coords,
            nodes near coords align={
                yshift=6.3
            },
            every node near coord/.style={
                font=\tiny,
                color=black
            }
        ] table [x=GPUs, y=one, col sep=comma] {figures/data/DDP_scaling_pgfplots.csv};
        \addlegendentry{1-way}
        \addplot[
            color=kit-blue, 
            mark=*, 
            mark options={
                scale=0.5, 
            },
            point meta=explicit symbolic, 
            nodes near coords,
            nodes near coords align={
                yshift=6.3
            },
            every node near coord/.style={
                font=\tiny,
                color=black
            }
        ] table [x=GPUs, y=two, col sep=comma] {figures/data/DDP_scaling_pgfplots.csv};
        \addlegendentry{2-way}
        \addplot[
            color=kit-orange, 
            mark=*, 
            mark options={
                scale=0.5, 
            },
            point meta=explicit symbolic, 
            nodes near coords,
            nodes near coords align={
                yshift=6.3
            },
            every node near coord/.style={
                font=\tiny,
                color=black
            }
        ] table [x=GPUs, y=four, col sep=comma] {figures/data/DDP_scaling_pgfplots.csv};
        \addlegendentry{4-way}
    \end{axis}
\end{tikzpicture}
    \caption{Efficiency for weak scaling experiments combining intra-node MP and inter-node DP.}
    \label{fig:DDP}
    \Description{This needs a description}
\end{figure}
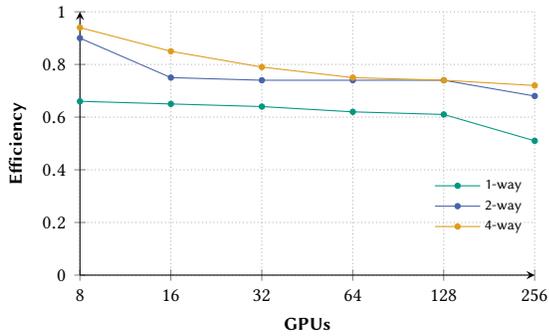

Figure~\ref{fig:DDP} shows weak scaling efficiency results.
At the largest scale, on 256 NVIDIA A100-40GB GPUs, we reach a total performance of 11 PFLOPs and 9 PFLOPs with the 2- and 4-way models, with mixed TF32 precision. This translates to 28\% and 23\% of theoretical peak performance, showing good scalability considering that this is a real-world application.
When considering the scaling efficiency relative to the baselines, we see that 2- and 4-way models scale far better than the native implementation, with final scaling efficiencies of 68\% and 72\%, respectively. The non-MP model has 51\% efficiency. This is due to the sharding of the model and optimizer states, meaning that communication is only required between the same model shards across nodes. These results imply that MP leads to higher scalability across large systems, compared to native single model instance scaling.
The large size of the input data---specifically, the large number of spatial tokens within the mixing blocks---leads to large communication volumes in the gradient reduction steps. These DP weak scaling results for  input data  with global resolutions of 0.25\degree{} have not yet been shown by other works in this domain.

\subsubsection{Energy usage}
In line with recent trends in AI research towards sustainable AI research~\cite{debus2023reporting}, we report the energy used for final training runs and scaling experiments, i.e., all experiments required to generate the roofline plot and the weak inter-node scaling experiments. The obtained energy usage measurements and corresponding CO\textsubscript{2} equivalents can be found in Table~\ref{tab:energy-usage}.  
CO\textsubscript{2} ($CO_2e$) equivalent emissions were estimated with $CO_2e = E_{total}  * PUE * e_C$, 
where $PUE$ is the power utilization effectiveness, the ratio between the energy used by compute components and the energy used for the entire data center infrastructure, equal to $1.05$ for HoreKa; and $e_C$ is the carbon efficiency, the ratio of carbon and equivalent gas emissions 
and equal to approximately \SI{381}{\gcoee}~\cite{UBA}. 
The total energy consumption of the three training runs and scaling experiments amounts to 2\,000 kWh, which corresponds to 2.5 months of the average U.S. household electricity consumption\footnote{\url{https://www.eia.gov/energyexplained/use-of-energy/electricity-use-in-homes.php}}.

\begin{table}[h!]
    \caption{Power draw for experiments as reported by the system's sensors. Scaling includes all roofline and DP runs.}
    \label{tab:energy-usage}
    \begin{tabular}{lccc}
        \toprule
        \multicolumn{1}{c}{\shortstack{\textbf{Experiment}\\{}}} & \multicolumn{1}{c}{\shortstack{\textbf{Energy usage} \\ \textbf{[kWh]}}} & \multicolumn{1}{c}{\shortstack{\textbf{CO\textsubscript{2} equivs.} \\\textbf{[kg]}}} & \multicolumn{1}{c}{\shortstack{\textbf{Total}\\ \textbf{[GPUh]}}}
        \\
        \midrule
        \textbf{1-way} & 579 & 232 & 1380 \\
        \textbf{2-way} & 643 & 257 & 1510 \\
        \textbf{4-way} & 855 & 342 & 2066 \\
        \midrule 
        \textbf{Scaling} & 445 & 178 & 1060 \\
        \bottomrule
    \end{tabular}
\end{table}

\section{Conclusion}
Current trends in data-driven atmospheric modeling, in particular considering large-scale foundation models, are developing towards higher spatial resolutions, longer rollout lengths, and increasing model capacity. However, these aspects pose significant computational challenges to existing neural model architectures and hardware.
To address these challenges, we introduce WeatherMixer and Jigsaw parallelism: two approaches towards scalable, high-resolution data-driven modeling of the atmosphere. 
With WeatherMixer, we present the first MLP-Mixer architecture in the field as a competitive alternative to dominating Transformer-based architectures. 
WeatherMixer scales linearly with respect to input sequence length, in comparison to Transformers, which scale quadratically. 
The simplicity of this MLP-based architecture offers great potential for model parallelism: Unlike the complex computational pathways in Transformer-based models, WeatherMixer reduces to a sequence of matrix--matrix multiplications. We exploit this with Jigsaw and introduce a competitive parallelization scheme that leverages domain, tensor, and data parallelism for efficient hardware utilization. 

Through parallelization with Jigsaw, and in particular domain-parallelism, WeatherMixer exhibits superscalar weak scaling in systems that are limited by data I/O-bandwidth---a problem frequently encountered in scientific applications of AI, where spatio-temporally resolved datasets are used for model training. 
Though future hardware improvements are likely to move towards larger memory bandwidth and faster performance through mixed-precision computing, we hypothesize that the input data size will grow faster than hardware developments, and developers will consistently be operating at the memory limits of their systems. Jigsaw thus offers an attractive solution on future hardware systems.

Furthermore, through its zero-redundancy memory partitioning, Jigsaw allows for large model and data sizes to be stored efficiently in memory. This allows us to scale WeatherMixer beyond the 1-billion-parameter mark without the need for elaborate sharding schemes used in other works. 
Not only does Jigsaw allow for larger models to be trained, but we also present empirical results that show that predictive performance benefits from tensor parallelism because the problem of large-batch effects is naturally mitigated. By assigning more GPUs of a fixed compute budget to model partitioning instead of data-parallelism, the model uses smaller global batch sizes and thus converges to lower loss values.

Next to these innovative contributions, our work also highlights the importance of analyzing scaling behavior of training large-scale AI models by means of high-performance computing paradigms. Through roofline analyses and weak and strong scaling experiments, we enhance understanding of the interplay and relative performances between data loading, data transfer, and computational workload, allowing for optimal deployment on HPC systems.

\begin{acks}
This research has been supported by the German Federal Ministry of Education and Research under the 01LK2313A SMARTWEATHER21-SCC-2 grant and by the KIT Center MathSEE under the FAST-DREAM Bridge PhD grant. This work was performed on the HoreKa supercomputer funded by the Ministry of Science, Research and the Arts of Baden-Württemberg and by the Federal Ministry of Research, Technology and Space. The authors gratefully acknowledge the computing time made available to them on HoreKa at NHR at KIT via the SmartWeather21-p0021348 NHR large project. This center is jointly supported by the Federal Ministry of Education, Technology and Space and the state governments participating in the NHR (https://www.nhr-verein.de/unsere-partner).
\end{acks}

\bibliographystyle{ACM-Reference-Format}
\bibliography{references}

\end{document}